\documentclass[journal]{IEEEtran}
\usepackage{amsmath,amsfonts}
\usepackage{algorithmic}
\usepackage{algorithm}
\usepackage{array}
\usepackage[caption=false,font=normalsize,labelfont=sf,textfont=sf]{subfig}
\usepackage{textcomp}
\usepackage{stfloats}
\usepackage{url}
\usepackage{verbatim}
\usepackage{graphicx}
\usepackage{xcolor}
\usepackage{cite}
\usepackage{multirow}

\begin{document}

\title{Weakly Supervised Realtime Dynamic Background Subtraction}

\author{Fateme Bahri and Nilanjan Ray\\
Department of Computing Science\\
University of Alberta, Canada\\
{\tt\small \{fbahri,nray1\}@ualberta.ca}
}

\maketitle

\begin{abstract}
Background subtraction is a fundamental task in computer vision with numerous real-world applications, ranging from object tracking to video surveillance. However, dynamic backgrounds can pose a significant challenge in this problem. While various methods have been proposed for background subtraction, supervised deep learning-based techniques are currently considered state-of-the-art. However, these methods require pixel-wise ground-truth labeling, which can be time-consuming and expensive. In this work, we propose a weakly supervised framework that can perform background subtraction without requiring per-pixel ground-truth labels. Our framework is trained on a moving object-free sequence of images and comprises two networks. The first network is an autoencoder that generates static background images and prepares dynamic background images for training the second network. The dynamic background images are obtained by thresholding the background-subtracted images. The second network is a U-Net that uses the same moving object-free video for training and the dynamic background images as pixel-wise ground-truth labels. During the test phase, the input images are processed by the autoencoder and U-Net, which generate static and dynamic background images, respectively. The dynamic background image helps remove dynamic motion from the background subtracted image, enabling us to obtain a foreground image that is free of dynamic artifacts. To demonstrate the effectiveness of our method, we conducted experiments on selected categories of the \textit{CDnet 2014} dataset and the \textit{I2R} dataset. Our method outperformed all top-ranked unsupervised methods. It also surpassed one of the two existing weakly supervised methods, while achieving comparable results to the other method but with a shorter running time. Our proposed method is online, realtime, efficient, and requires minimal frame-level annotation, making it suitable for a wide range of real-world applications.
\end{abstract}

\section{Introduction} \label{sec:Introduction}
Background subtraction is a crucial problem in computer vision that has practical applications in various domains like video surveillance, human-computer interaction, traffic monitoring, and autonomous navigation \cite{bouwmans2019deep, garcia2020background}. Dealing with dynamic backgrounds is a significant challenge in background subtraction, where a background pixel's value can change due to periodical or irregular movements \cite{xu2016background}. Although various methods have been proposed for background subtraction, not all of them can effectively handle sequences with dynamic backgrounds. Scenes with dynamic elements like fountains, waving trees, and water motions are prime examples of dynamic backgrounds. Detecting these dynamic variations as parts of the foreground negatively impacts the performance of the methods.

Statistical methods are among the simplest approaches to dealing with dynamic backgrounds. These methods utilize statistical modeling of pixel value distributions. Examples include the Gaussian mixture model (GMM) \cite{stauffer1999adaptive} and its improved variations \cite{zivkovic2004improved, zivkovic2006efficient, lee2005effective, kaewtrakulpong2002improved}, as well as kernel density estimation (KDE) \cite{elgammal2000non}.
 
Another class of methods involves dynamically adjusting their parameters through feedback mechanisms. The SuBSENSE method \cite{st2014subsense} was the pioneering method in this category, and it has inspired several other methods, including PAWCS \cite{st2015self}, SWCD \cite{isik2018swcd}, and CVABS \cite{icsik2019cvabs}.

Because of the effectiveness of deep learning methods in computer vision, numerous neural network models have been developed for the purpose of foreground and background segmentation. These models have ranked highly among the evaluated methods in the \textit{CDnet 2014} dataset. However, they require supervised training, which involves manual annotation at the pixel level. This process is time-consuming and expensive, and may not be practical for every situation.

In recent years, weakly supervised methods have gained popularity and have demonstrated impressive performance in various tasks. One of their primary advantages is that they achieve satisfactory performance without relying on costly pixel-wise ground truth annotations. In the context of background subtraction, two new methods have been proposed by Zhang et al. \cite{zhang2022learning} and Minematsu et al. \cite{minematsu2019simple}. Both methods are trained using frame-level labels, which is a less demanding and more cost-effective labeling approach compared to pixel-level annotation.

We present a new approach to background subtraction that learns the dynamic background component in a weakly supervised manner. It uses a fully connected autoencoder and a U-Net convolutional neural network \cite{ronneberger2015u}. To explain the overall working principle, let us first consider the scenario where no moving object is present in a sequence. In this case, the autoencoder takes in the sequence and produces the static background images. The difference between the input image and the output of the autoencoder contains dynamic clutter. The U-Net takes in the same sequence of images and is expected to produce only the dynamic clutter, which in this paper is referred to as the dynamic background image. So, when we subtract the autoencoder output from the input image and multiply it with the inverted output of the U-Net, we ideally obtain a zero image showing no moving objects or dynamic background. In the second scenario, when moving objects are present, the autoencoder output again contains static background and the output of the U-Net is still expected to produce only a dynamic background image. So, when we subtract the output of the autoencoder from the input image and multiply it with the inverted output of the U-Net, it will show the moving objects only.

The autoencoder is trained on a moving object-free sequence to produce static background images that capture some of the temporal and spatial variations in the scene. We obtain a binary dynamic background image by subtracting autoencoder output from the input image and applying a threshold. Then, We train the U-Net on the same object-free sequence using the binary dynamic background images from the previous step as the target output for the U-Net. Thus, the U-Net learns the temporal and spatial variations of the dynamic background in the scene. 

\begin{figure*}[htbp]
\centering
\includegraphics[width=\textwidth]{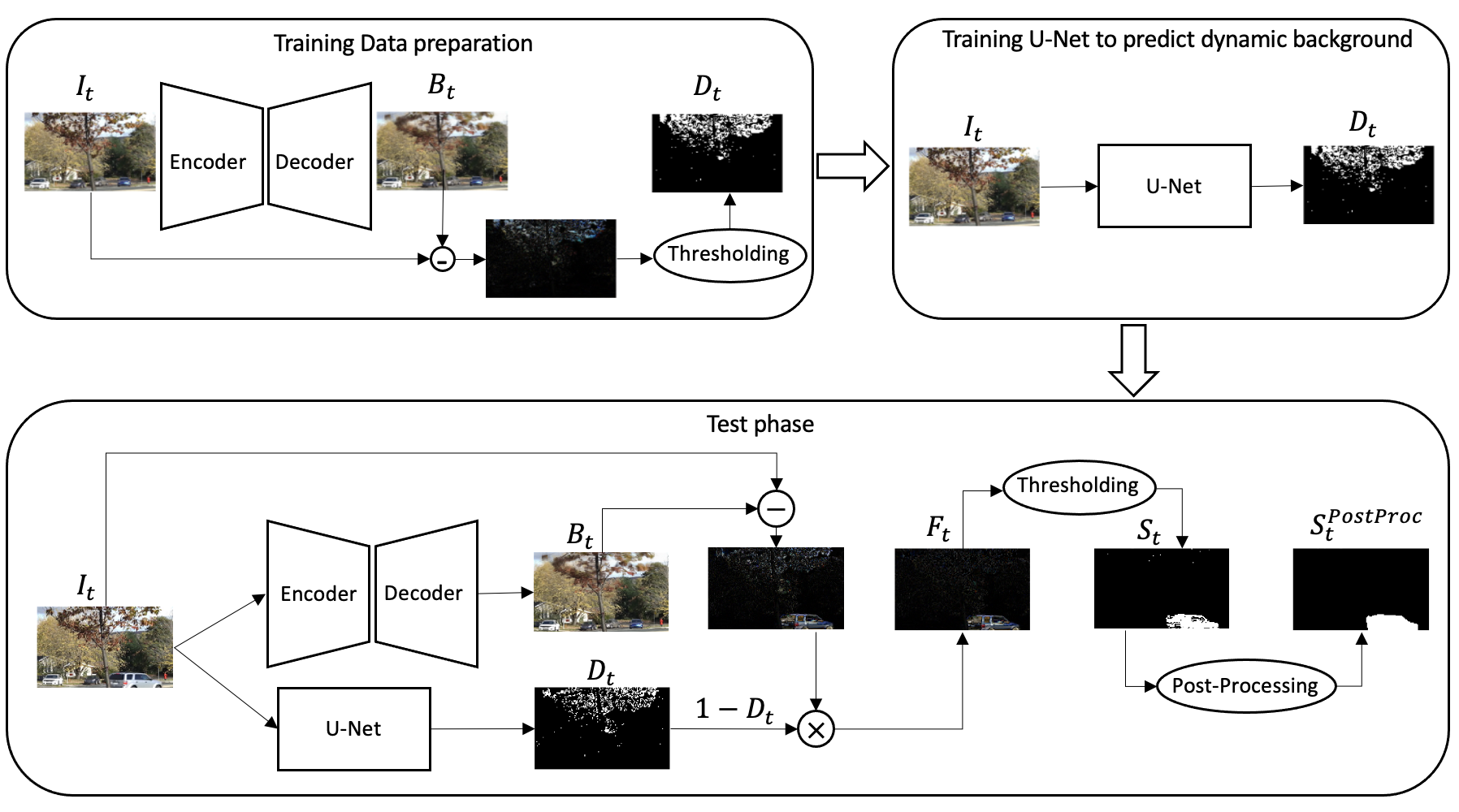}
\caption{Overview of the proposed method. The top two boxes show the training phase and the below box shows the test phase. To handle dynamic background, our method combines an autoencoder and a U-Net. The autoencoder and U-Net are trained to generate the static and dynamic background images, respectively. By subtracting the autoencoder's output from the input image and multiplying it with the inverted output of the U-Net, we can obtain an image that shows only the moving objects, free from dynamic artifacts.}
\label{fig:overview}
\end{figure*}

During the testing phase, the autoencoder and U-Net generate the static and dynamic background images, respectively. Multiplying the inverted dynamic background with the static background-subtracted image produces the foreground image. Finally, we apply pixel-wise thresholding to the foreground image and use some simple post-processing techniques to enhance the final result. Fig. \ref{fig:overview} illustrates an overview of our proposed method.

Our proposed approach offers several key contributions. First, it is a practical and cost-effective weakly supervised framework, which eliminates the need for costly pixel-wise annotations by using only an object-free training sequence. Second, it effectively learns and predicts the dynamic background component in each image and segments it from the foreground. Finally, our experimental results show that the proposed algorithm achieves superior performance in dynamic background scenes compared to other state-of-the-art unsupervised and weakly supervised methods, both quantitatively and qualitatively.

The paper is structured as follows: Section \ref{sec:Related work} reviews other background subtraction methods. Section \ref{sec:Method} provides a detailed explanation of our framework, including its training and test phases. In Section \ref{sec:Results}, we present our implementation details, and experimental results, and compare them with state-of-the-art methods. Finally, Section \ref{sec:Conclusion} concludes the paper with a summary of our findings and future research directions.

\section{Related Works} \label{sec:Related work}
\subsection{Statistical Methods}
There is a category of methods that use statistical approaches based on probability density estimation of pixel values. The most basic of these is the single Gaussian model \cite{wren1997pfinder}. However, this approach has limitations as a single function cannot account for all variations in pixel values. To overcome this, the Gaussian mixture model (GMM) \cite{stauffer1999adaptive} was proposed, which uses several Gaussian densities. Various improved versions of this traditional and widely used method have been presented \cite{zivkovic2004improved, zivkovic2006efficient, lee2005effective, kaewtrakulpong2002improved} with better results. Flux Tensor with Split Gaussian models (FTSG) \cite{wang2014static} is a state-of-the-art method that uses flux tensor-based motion segmentation and GMM-based background modeling separately and then merges the results. Finally, it enhances the results using a multi-cue appearance comparison. However, parametric methods such as GMM and its successors are unable to handle sudden changes in a scene. To address this issue, a statistical non-parametric algorithm called KDE \cite{elgammal2000non} was introduced, which uses kernel density estimation to model the probability of pixel values.

\subsection{Methods Based on Dynamic Feedback Mechanism}
One of the main categories of methods for background modeling involves using controller parameters that update the background model based on dynamic feedback mechanisms. One such method, SuBSENSE \cite{st2014subsense}, incorporates color channel intensity and spatiotemporal binary features and adjusts its parameters using pixel-wise feedback loops based on segmentation noise. PAWCS \cite{st2015self}, a newer and more advanced method, extends the capabilities of SuBSENSE by generating a strong and persistent dictionary model based on spatiotemporal features and color. Similar to SuBSENSE, PAWCS also employs automatic feedback mechanisms to adjust its parameters. Another method, SWCD \cite{isik2018swcd}, combines the dynamic controllers of SuBSENSE with a sliding window approach to update background frames. Finally, CVABS \cite{icsik2019cvabs}, a recent subspace-based method, utilizes dynamic self-adjustment mechanisms like SuBSENSE and PAWCS to update the background model.

\subsection{Ensemble Methods}
Ensemble methods have emerged as a new approach for change detection algorithms. The authors of \cite{bianco2017combination,bianco2017far} have recently introduced a method named IUTIS (In Unity There Is Strength) that utilizes genetic programming (GP) to combine different algorithms and maximize their individual strengths. By selecting the best methods, combining them in various ways, and applying appropriate post-processing techniques, GP enables IUTIS to achieve high performance. The method shows promising performance by integrating several top-ranked methods evaluated on \textit{CDnet 2014} (\cite{wang2014cdnet}).

\subsection{Deep Learning Methods}
Several deep neural networks (NN) have been proposed in recent years for foreground segmentation, owing to the success of deep learning in computer vision. FgSegNet and its variations \cite{gao2021extracting,lim2018foreground,lim2020learning} are presently considered state-of-the-art based on their performance on \textit{CDnet 2014}. Motion U-Net \cite{rahmon2021motion} is another deep NN method that requires fewer parameters than FgSegNet. BSPVGAN \cite{zheng2020novel} employs Bayesian Generative Adversarial Networks (GANs) to create a background subtraction model. Another technique called Cascade CNN \cite{wang2017interactive} uses a multi-resolution convolutional neural network (CNN) to segment moving objects. In DeepBS \cite{babaee2018deep}, a CNN is trained using patches of input images, which are then merged to reconstruct the frame. Temporal and spatial median filtering is utilized to enhance the segmentation outcomes. Another supervised approach, BSUV-Net \cite{tezcan2020bsuv,tezcan2021bsuv}, is trained on some image sequences along with their spatiotemporal data augmentations, and exhibits good performance on unseen videos after training. Among the assessed methods on \textit{CDnet 2014}, the aforementioned neural network techniques are ranked at the top. However, they require supervised training, which entails pixel-wise annotated ground truth, a time-consuming and impractical task in many situations.

A number of recently developed techniques, including SemanticBGS \cite{braham2017semantic} and its variations RT-SBS-v1 and RT-SBS-v2 \cite{cioppa2020real}, integrate semantic segmentation with background subtraction methods. They employ the information from a semantic segmentation algorithm to obtain a pixel-wise probability that enhances the output of any background subtraction method. However, we cannot compare them to our method because they rely on pixel-level information as input, even though they are not trained using ground-truth labels.

\subsection{Weakly Supervised Methods}
In recent years, some weakly supervised methods have emerged that solely rely on image-level tags, which indicate whether a foreground object is present in the image \cite{zhang2022learning, minematsu2019simple}. The method proposed in \cite{minematsu2019simple} generates a binary mask image by subtracting a background image from an input image to identify foreground regions. It then uses intermediate feature maps of a CNN to refine the foreground locations. However, image-level supervision presents a challenge due to the lack of location information in training the network. To address this, the method introduces some constraints that help to locate foreground pixels.  

Another recent technique, LDB \cite{zhang2022learning}, adopts a tensor-based decomposition framework to represent the background as a low-rank tensor and classify the sparse noise as foreground. Additionally, it trains a two-stream neural network using an object-free video to explicitly learn the dynamic background. The dynamic background component of LDB leads to a more precise decomposition of the background and foreground, making it the current state-of-the-art method in weakly supervised moving object detection, particularly in dynamic background scenes.

Our proposed method is also based on explicit modeling of the dynamic background using a neural network. However, our framework differs from LDB in that it relies exclusively on neural networks for the segmentation of the background and foreground, rather than using a low-rank-based approach. As a result, we gain significant advantages in reducing the running time once our networks are trained. Further, LDB uses a very light CNN to model dynamic background. Instead, we use a U-Net to model the same. Because of a large number of parameters, U-Net has significant representative power, and it \textbf{overfits} the scene sequence to generate the dynamic background. We make use of this overfitting, because for a sequence containing moving objects the U-Net should ignore the moving objects and output only the dynamic background.

\section{Proposed Method} \label{sec:Method}
Our proposed method, depicted in Fig. \ref{fig:overview}, is based on two neural networks and is trained in a weakly supervised way. The first network is an autoencoder that generates static background images and is trained in an unsupervised manner. The second network is a U-Net \cite{ronneberger2015u}, which requires pixel-wise labels for training. Using the background images generated by the autoencoder, ground-truth labels for the U-Net are acquired. In the following sections, we explain the training and testing phases, as well as the roles of each network in our framework.

\subsection{Background Generation}
Our framework uses an autoencoder to generate static background images. Autoencoders are a type of neural network that consists of two components: an encoder and a decoder. The encoder maps the input to a compressed code, and the decoder reconstructs the input from the code, aiming to make the output as close to the input as possible \cite{bank2020autoencoders}. Consequently, autoencoders learn a compressed and meaningful representation of the input data. This results in the removal of insignificant data and noise from the reconstructed input \cite{vincent2008extracting}.

\begin{figure*}[htbp]
\centering
\includegraphics[width=\textwidth]{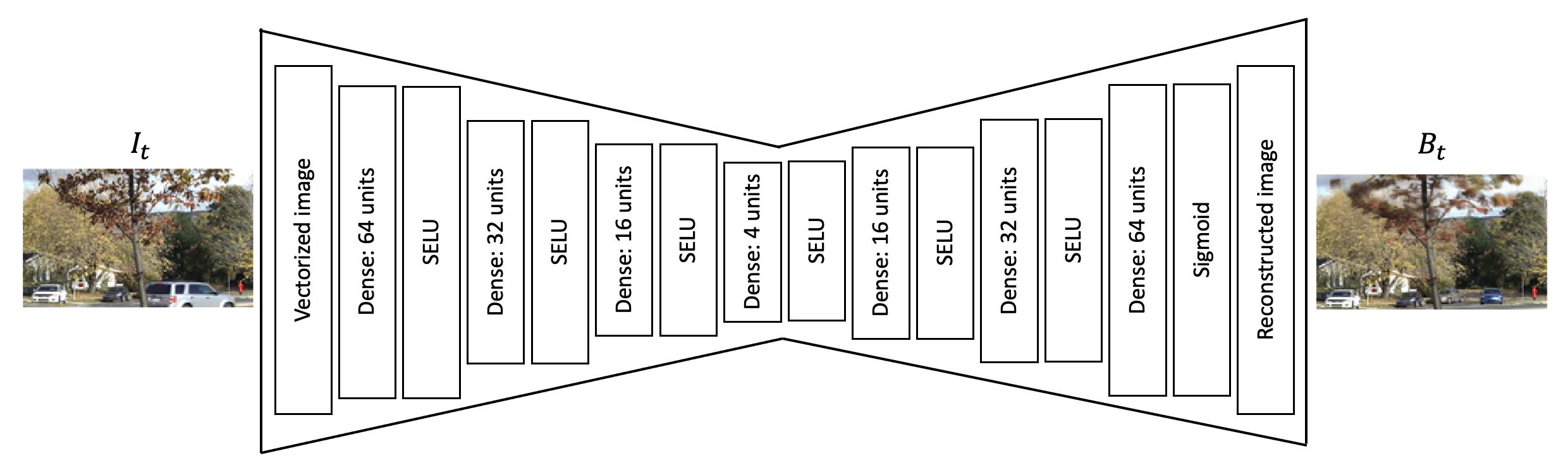}
\caption{The autoencoder architecture has fully connected layers, each followed by a SELU activation function. The last layer has a Sigmoid activation function to limit the output values between zero and one.}
\label{fig:autoencoder}
\end{figure*}

The autoencoder used in our method for generating the static background images is a fully connected one with dense layers, each followed by a SELU \cite{klambauer2017self} activation function. The only exception is the last layer, which uses a Sigmoid activation function to limit the output values between zero and one. Fig. \ref{fig:autoencoder} shows the details of the architecture of the autoencoder used in our approach.

The $L_{recons}$ loss term, which is responsible for constructing the background images, is defined as follows (Eq. \ref{loss_Reconst}):
\begin{equation} \label{loss_Reconst}
L_{recons} =\sum_{t=1}^N\Vert I_t - B_t \Vert_1 ,
\end{equation}
Here, $I_t$ and $B_t$ are the $t^{\text{th}}$ input and output of the autoencoder, respectively, and $\Vert.\Vert_1$ denotes the $L_1$-norm. We used the $L_1$-norm instead of the $L_2$-norm in $L_{recons}$ because it encourages sparsity \cite{candes2008enhancing}.

The autoencoder can learn a low-dimensional manifold of the data distribution by applying constraints such as limiting the network's capacity and choosing a small code size ~\cite{goodfellow2016deep}. The network can extract the most salient features of the data, and the $L_{recons}$ loss term imposes similarity between the input and reconstructed frames, allowing the autoencoder to learn a background model during optimization. This is possible because the input image sequence is temporally correlated, and the background of the images is common among them \cite{bahri2018online}.

\subsection{Dynamic Background Data Preparation}
Autoencoders can be optimized in an unsupervised way and do not require labeled data. However, our second network, U-Net, requires pixel-wise labeled training data to be trained. A moving object-free sequence of input frames is used as training data. During the training phase, the autoencoder generates static background images of the training frames. The static background images are then subtracted from the input images, and after applying a threshold on them, the binary images are obtained. These binary images exhibit a dynamic background since they are extracted from training images without any foreground object. The entire process is illustrated in Fig. \ref{fig:overview}.

\subsection{Dynamic Background Prediction}
The second network in our proposed method is a U-Net, which was originally developed for image segmentation tasks and has shown great success in medical image analysis \cite{ronneberger2015u}. Its architecture resembles a U-shape and consists of two paths: a contracting path and an expansive path. The contracting path is composed of convolutional layers followed by ReLU activation functions and max-pooling layers, where the number of features gets doubled in each contracting step. The expansive path consists of up-convolutional layers for upsampling and halving the number of features, concatenations of features from the contracting path, convolutional layers, and ReLU activation functions. In our method, we used the basic U-Net architecture as described in \cite{ronneberger2015u}.

U-Net is a network that requires pixel-wise labels for supervised training. However, in our framework, it is trained using the prepared binary images that were explained in the previous section, which makes our method weakly supervised. All we need for training is a moving object-free sequence. In other words, if we have a training sequence with frame-level tags whether the frame contains only background or not, we select only the frames with a tag value of zero as the training data. The output of U-Net is a binary image with pixel values of zero or one, where each pixel labeled as one indicates the presence of dynamic background.

\subsection{Training and Test Phases}
\subsubsection{Training Phase}
In the training phase, our proposed method first optimizes the autoencoder on an object-free training sequence to generate static background images. Then, a threshold is applied to the background-subtracted images to obtain dynamic background binary images. The U-Net network is then trained on the same object-free training sequence using the dynamic background binary images as its target output to enable it to predict dynamic background pixels. We train the U-Net model long enough until it overfits to the training sequence. We exploit this overfitting to our advantage because during testing, the U-Net should identify only the dynamic background pixels as label one, while ignoring the moving objects present in the video sequence.  All the steps of the training phase are illustrated in Fig. \ref{fig:overview}.

\subsubsection{Test Phase}
During the test phase, an input test image $I_t$ is fed into the autoencoder to obtain the static background image $B_t$. The same input image $I_t$ is also fed into the U-Net, which produces a dynamic background image $D_t$. Next, the background subtracted image, $I_t-B_t$, is multiplied by the inverted dynamic background, $1-D_t$, as shown in equation \ref{eq_F}. This step generates the foreground image $F_t$, which excludes the dynamic background artifacts. Then, a pixel-wise thresholding technique is applied to $F_t$ to obtain the initial segmented image, $S_t$, as described in the next section. Finally, two standard post-processing techniques, median blurring, and morphological closing are applied to $S_t$ to improve the results, and the final segmented image, $S_t^{PostProc}$, is obtained. The entire process is illustrated in Fig. \ref{fig:overview}.

\begin{equation} \label{eq_F}
F_t = (I_t-B_t)\times(1-D_t)
\end{equation}

\subsection{Foreground Segmentation with Pixel-wise thresholding}
Although most of the dynamic background pixels are detected by the U-Net, there is still a possibility that some of them may be missed due to the selected threshold when preparing the dynamic background ground-truth images. To address this, we use a per-pixel thresholding technique inspired by the SuBSENSE method \cite{st2014subsense} to obtain the foreground masks. This technique calculates the dynamic entropy of each pixel, represented by the dynamic entropy map $C(x)$, to detect blinking pixels. The dynamic entropy map tracks how often a pixel changes from being a foreground pixel to a background pixel, or vice versa, between consecutive frames.

The calculation of $C(x)$ is based on the XOR operator and is given by:

\begin{equation} \label{eq_counter}
C(x) =\frac{1}{N-1}\sum_{t=2}^N XOR (S_t^{init}(x),S_{t-1}^{init}(x)),
\end{equation}

Here, $x$ represents a pixel, $S_t^{init}$ is the binary result of the $t^{\text{th}}$ frame in the sequence after an initial segmentation, and $N$ is the total number of frames in the sequence. The initial segmentation uses $\alpha \max(F)$ as the threshold, $\max(F)$ is the maximum value of the foreground frames $F$, and $\alpha$ is a coefficient. The dynamic entropy map values, $C$, range from 0 to 1.

In the next step, we compute the pixel-wise distance thresholds using the following equation:

\begin{equation} \label{eq_thresholds}
R(x) = \beta\max(F)+C(x),
\end{equation}

Here, $\max(F)$ is the maximum value of the foreground frames $F$, and $\beta$ is a coefficient. The binary segmented result $S_t$ is obtained by applying the distance threshold $R(x)$ to the foreground $F_t(x)$.

\section{Experimental Results and Discussion}\label{sec:Results}

\subsection{implementation details}
 Our method was implemented using the Keras deep learning framework \cite{chollet2015keras}. The autoencoder architecture shown in Figure \ref{fig:autoencoder} consists of densely connected layers with 64, 32, 16, 4, 16, 32, and 64 units, respectively. All layers use the scaled exponential linear unit (SELU) activation function \cite{klambauer2017self}, except for the final layer which uses the sigmoid activation function to produce output values within the range of $[0,1]$.

The U-Net's contracting path consists of four steps, each composed of two convolutional layers with a $3\times3$ kernel size and ReLU activation function, followed by a $2\times2$ max pooling operation with a stride of 2. The numbers of features are 64, 128, 256, 512, and 1024 for the top-to-bottom steps. The expansive path mirrors the contracting path but with two differences: first, features of the same contracting level are concatenated to the feature channels. Second, the max pooling operation is replaced with a transposed convolution layer with a $3\times3$ kernel size and stride 2. Consequently, the number of features is halved in each expansive step. The final layer of the model is a convolutional layer with a $1\times1$ kernel size and two features that construct the binary output image. Our U-Net architecture has the same design as the basic U-Net proposed in \cite{ronneberger2015u}, except that we use convolutional and transposed convolution layers with the padding mode set to 'same', which eliminates the need for the cropping operation in \cite{ronneberger2015u}.

The hyper-parameters $\alpha$ and $\beta$ were set to $0.2$ and $0.08$, respectively, after conducting several trial and error experiments. We used the Adam optimization algorithm \cite{kingma2014adam} with learning rates of $0.0001$ and $0.005$ for the autoencoder and U-Net, respectively. Both networks were trained for 50 epochs. During the testing phase, we achieved an average processing speed of 107 frames per second on the \textit{CDnet 2014} dataset \cite{wang2014cdnet} using a GeForce GTX 1080 Ti GPU.

\subsection{Datasets}
We evaluated the effectiveness of our approach on various video datasets to demonstrate its suitability for real-world scenarios.

\subsubsection{CDnet 2014}
To show the effectiveness of our method in challenging dynamic background scenarios, we conducted evaluations on the \textit{Dynamic Background} category of the \textit{CDnet 2014} dataset \cite{wang2014cdnet}. This category consists of six videos with different types of dynamic background motions. The videos ``fountain01'' and ``fountain02'' feature a dynamic water background, while ``canoe'' and ``boats'' depict water surface movement. ``Overpass'' and ``fall'' exhibit waving trees in the background. Additionally, we evaluated our approach on the \textit{Bad Weather} category of the same dataset, which features sparse dynamic variations in the background caused by snow and rain, making it a challenging category. The four videos in this category are ``blizzard'', ``skating'', ``snowFall'', and ``wetSnow''.

We manually selected the frames without objects for the training data for each sequence. These frames were chosen from the frames before the starting frame in the temporal ROI. In the case of the ``WetSnow'' video, no background images were available in the initial frames of the sequence. Therefore, we chose 10 object-free frames from the sequence after frame number 2000. We used a maximum of 300 frames for training, or the number of available frames, whichever was less.

\subsubsection{I2R Dataset}
The \textit{I2R} dataset \cite{li2004statistical} is a widely recognized benchmark for background subtraction tasks, consisting of 10 real videos shot in indoor and outdoor settings. These videos contain challenging conditions like bootstrapping, shadows, camouflage, lighting changes, noise, weather, and dynamic backgrounds. To assess our approach, we chose three outdoor scenes with dynamic backgrounds: ``Campus,'' ``Fountain,'' and ``WaterSurface.'' As outlined in the previous section, we manually selected the training frames.

\subsection{Evaluation Metric}
To evaluate the performance of our method, we utilize the F-Measure (FM) metric, which is a widely used performance indicator for moving object detection and background subtraction algorithms. The F-measure is calculated using the equation shown in (\ref{F-Measure}), which combines the recall and precision scores. In order to maintain consistency with existing methods, we compute all evaluation metrics according to the definitions provided in \cite{wang2014cdnet}.

\begin{equation} \label{F-Measure}
\text{F-measure}  = 2* \frac {\text{Recall}*\text{Precision}}{\text{Recall}+\text{Precision}}
\end{equation}

\subsection{Qualitative Results}
\begin{figure*}[htbp]
\centering
\includegraphics[width=\textwidth]{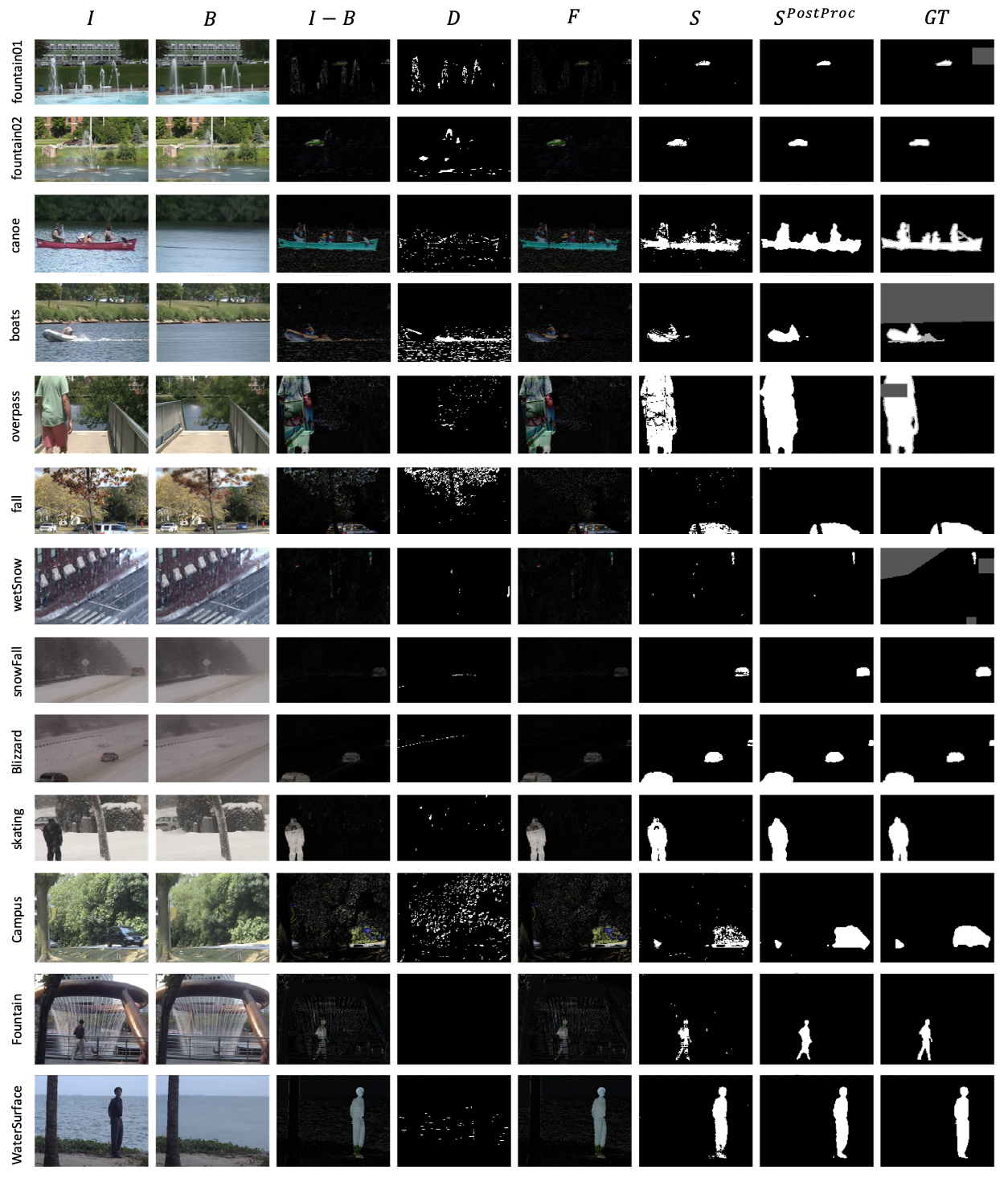}
\caption{Qualitative results of the proposed method. The columns from left to right display the input images, the generated background images, the background-subtracted images, the predicted dynamic background images, the obtained foreground images, the initial segmented images after thresholding, the final segmented images after post-processing and the ground-truth images, respectively}
\label{fig:qualitative}
\end{figure*}

In Fig. \ref{fig:qualitative}, we present the intermediate and final qualitative results of our method's steps on the videos. The first six rows depict the \textit{Dynamic Background} category, followed by the next four rows from the \textit{Bad Weather} category of the \textit{CDnet 2014} dataset. The last three rows show videos from the \textit{I2R} dataset.

The first three columns display the input frame, the autoencoder-generated background, and the background-subtracted image, respectively. The fourth column exhibits the dynamic background image predicted by the U-Net. The next column displays the foreground image obtained by multiplying the background-subtracted image with the inverted version of the dynamic background image. The sixth and seventh rows showcase the initial segmented image after thresholding and the final segmented image after post-processing, respectively. The last column shows the ground-truth images.

In Fig. \ref{fig:qualitative}, it is evident from the 4th column that the U-Net model can efficiently capture the dynamic background motion and create a precise representation of the dynamic background image, particularly for the \textit{Dynamic Background} videos. By comparing the background-subtracted image in the third column with the obtained foreground in the fifth column, it proves our method can effectively decompose the dynamic background from the foreground objects.

For the \textit{Bad Weather} videos, our method is able to predict some of the dynamic background pixels. However, due to the nature of our autoencoder, it tends to absorb snow noise in the generated static background image, leading to a lack of visible snow pixels in the dynamic background image. Nevertheless, our method is still able to produce high-quality results, demonstrating its effectiveness in handling challenging weather scenarios.

Regarding the \textit{I2R} dataset, our U-Net accurately predicts the dynamic background pixels in the ``Campus'' sequence and some of the dynamic background pixels in the ``WaterSurface'' sequence. However, in the ``Fountain'' sequence, the U-Net is not able to predict the fountain pixels in the dynamic background image since they are already detected as part of the background generated by the autoencoder. This is because the values of the fountain pixels remain constant in consecutive frames, and therefore, they are absorbed in the static background image.

The key aspect is to effectively separate the foreground pixels from the dynamic background pixels, which our framework achieves well through the use of both the autoencoder and the U-Net models. This ultimately leads to the superior performance of our method.

 \subsection{Quantitative Results}
In this section, we present the quantitative results of our method compared to the top-performing methods on the \textit{CDnet 2014} dataset \cite{wang2014cdnet} listed on ChangeDetection.net website. Specifically, we chose the top 30 methods based on their average F-measure (FM) performance on the \textit{Dynamic Background} and \textit{Bad Weather} categories, excluding supervised methods and the ensemble method IUTIS \cite{bianco2017combination}. We also included the results of the LDB weakly supervised method \cite{zhang2022learning}, as well as the CANDID algorithm \cite{mandal2018candid}, which was specifically proposed for dynamic background subtraction.

The results are shown in Table \ref{table:quantitative_results}. The second to eighth columns display the results on \textit{Dynamic Background} videos, and the ninth to thirteenth columns show the results on \textit{Bad Weather} videos. The methods are sorted based on their average FM on \textit{Dynamic Background} videos, which is listed in the eighth column. Our method's results are reported in the last row of the table.

As shown in Table \ref{table:quantitative_results}, our method achieves an average FM of 0.91 on \textit{Dynamic Background}, outperforming all unsupervised methods and the LDB \cite{zhang2022learning} method. Our method also achieves the highest FM on the ``fall'' video and performs the best on ``fountain01'' along with the FTSG method \cite{wang2014static}. These results demonstrate the practicality of our method for dynamic background scenes with only the cost of frame-level tag training data.

On \textit{Bad Weather} sequences, our method achieves an average FM of 0.89, outperforming all unsupervised top-ranked methods, but is surpassed by the LDB weakly supervised method \cite{zhang2022learning}, which has an FM of 0.91. Our method also achieves the best FM on the ``blizzard'' video, while LDB obtains the best FM on the ``WetSnow'' and ``snowFall'' sequences.

To compare our method to LDB more comprehensively, we also perform experiments on the \textit{I2R} dataset and report the results in Table \ref{table:I2R_results}. As shown, our method achieves the same average FM as LDB, but we obtain slightly better results on the ``Fountain'' and ``WaterSurface'' sequences, while LDB performs slightly better on the ``Campus'' sequence.

A comparison of our method and LDB on various videos shows that our method performs better in the \textit{Dynamic Background} category, while LDB performs better in the \textit{Bad Weather} category. For the \textit{I2R} sequences, both methods achieve similar performance. It is worth noting that LDB uses a low-rank tensor decomposition and is a batch method, whereas our method is an online method achieving realtime performance where an input image is fed through the two networks to obtain the result. 

We also compared our method to another weakly supervised method \cite{minematsu2019simple} described in the related work section. Minematsu et al. performed experiments on eight categories of the \textit{CDnet 2014} dataset \cite{wang2014cdnet}, but only selected some of the sequences for each category. We performed the same experiments and report the results in Table \ref{table:comparasion2weakly2}. As shown in the table, our method achieves an average FM of $0.72$, while Minematsu et al. \cite{minematsu2019simple} obtain an average FM of $0.66$. Our method outperforms theirs on the \textit{Bad Weather}, \textit{Dynamic Background}, \textit{Shadow}, \textit{Thermal}, and \textit{Turbulence} sequences, while they achieve better performance on the \textit{Baseline}, \textit{Camera Jitter}, and \textit{Night Videos} sequences.

\begin{table*}[ht!]
\caption{Performance comparison of the top-ranked methods, evaluated on \textit{CDnet 2014} \textit{Dynamic Background} and \textit{bad Weather} categories, in terms of F-measure. The best performance achieved, in each column, is shown in bold. Methods are sorted based on their Average F-Measure on \textit{Dynamic Background} Category}
\centering
\label{table:quantitative_results}
\begin{tabular}{l|ccccccc|ccccc}
\hline
\multirow{2}{*}{Methods} & \multicolumn{7}{c|}{Dynamic Background}                   & \multicolumn{5}{c}{Bad Weather}               \\ \cline{2-13} 
                         & fount01 & fount02 & canoe & boats & overpass & fall & Avg & wetSnow & snowFall & blizzard & skating & Avg \\ \hline
GraphCutDiff \cite{miron2015change}                                    & 0.08                           & 0.91                           & 0.57                      & 0.12                      & 0.84                         & 0.72                     & 0.54                         & 0.83                        & 0.90                         & 0.86                         & 0.92                        & 0.88                        \\
CL-VID \cite{lopez2018foreground}                                      & 0.05                           & 0.45                           & 0.93                      & 0.81                      & 0.85                         & 0.23                     & 0.55                         & 0.54                        & 0.79                         & 0.75                         & 0.87                        & 0.74                        \\
C-EFIC \cite{allebosch2015c}                                           & 0.27                           & 0.34                           & 0.93                      & 0.37                      & 0.90                         & 0.56                     & 0.56                         & 0.65                        & 0.74                         & 0.86                         & 0.90                        & 0.79                        \\
EFIC \cite{allebosch2015efic}                                          & 0.23                           & 0.91                           & 0.36                      & 0.36                      & 0.88                         & 0.72                     & 0.58                         & 0.62                        & 0.71                         & 0.86                         & 0.92                        & 0.78                        \\
Multi\_ST\_BG \cite{lu2014multiscale}                                  & 0.14                           & 0.82                           & 0.48                      & 0.89                      & 0.84                         & 0.41                     & 0.60                         & 0.53                        & 0.71                         & 0.71                         & 0.59                        & 0.64                        \\
KDE-ElGamm \cite{elgammal2000non}                                      & 0.11                           & 0.82                           & 0.88                      & 0.63                      & 0.82                         & 0.31                     & 0.60                         & 0.57                        & 0.78                         & 0.77                         & 0.91                        & 0.76                        \\
CP3-online \cite{liang2015co}                                          & 0.54                           & 0.91                           & 0.63                      & 0.17                      & 0.64                         & 0.77                     & 0.61                         & 0.75                        & 0.76                         & 0.85                         & 0.63                        & 0.75                        \\
DCB \cite{krungkaew2016foreground}                                     & 0.40                           & 0.83                           & 0.45                      & 0.87                      & 0.83                         & 0.30                     & 0.61                         & 0.30                        & 0.34                         & 0.41                         & 0.49                        & 0.38                        \\
GMM\_Zivk \cite{zivkovic2004improved}                                  & 0.08                           & 0.79                           & 0.89                      & 0.75                      & 0.87                         & 0.42                     & 0.63                         & 0.58                        & 0.76                         & 0.86                         & 0.76                        & 0.74                        \\
GMM\_Grim \cite{stauffer1999adaptive}                                  & 0.08                           & 0.80                           & 0.88                      & 0.73                      & 0.87                         & 0.44                     & 0.63                         & 0.61                        & 0.73                         & 0.88                         & 0.74                        & 0.74                        \\
SOBS\_CF \cite{maddalena2010fuzzy}                                     & 0.11                           & 0.83                           & \textbf{0.95}             & 0.91                      & 0.85                         & 0.26                     & 0.65                         & 0.50                        & 0.62                         & 0.67                         & 0.76                        & 0.64                        \\
SC\_SOBS \cite{maddalena2012sobs}                                      & 0.12                           & 0.89                           & \textbf{0.95}             & 0.90                      & 0.88                         & 0.28                     & 0.67                         & 0.50                        & 0.60                         & 0.66                         & 0.90                        & 0.66                        \\
AAPSA \cite{ramirez2016auto}                                           & 0.44                           & 0.36                           & 0.89                      & 0.76                      & 0.82                         & 0.75                     & 0.67                         & 0.63                        & 0.80                         & 0.82                         & 0.85                        & 0.77                        \\
M4CD\_V2 \cite{wang2018m}                                              & 0.17                           & 0.93                           & 0.61                      & \textbf{0.95}             & 0.95                         & 0.50                     & 0.69                         & 0.69                        & 0.81                         & 0.81                         & \textbf{0.94}               & 0.81                        \\
RMoG \cite{varadarajan2013spatial}                                     & 0.20                           & 0.87                           & 0.94                      & 0.83                      & 0.90                         & 0.67                     & 0.74                         & 0.60                        & 0.58                         & 0.76                         & 0.79                        & 0.68                        \\
WeSamBE \cite{jiang2017wesambe}                                        & 0.73                           & 0.94                           & 0.61                      & 0.64                      & 0.72                         & 0.81                     & 0.74                         & 0.81                        & 0.87                         & 0.90                         & 0.86                        & 0.86                        \\
Spectral360 \cite{sedky2014spectral}                                   & 0.47                           & 0.92                           & 0.88                      & 0.69                      & 0.81                         & 0.90                     & 0.78                         & 0.65                        & 0.79                         & 0.67                         & 0.92                        & 0.76                        \\
LDB Weak-Supervised \cite{zhang2022learning} & 0.14                           & 0.93                           & 0.92                      & 0.92                      & 0.95                         & 0.79                     & 0.78                         & \textbf{0.89}               & \textbf{0.93}                & 0.90                         & 0.90                        & \textbf{0.91}               \\
MBS\_V0 \cite{sajid2015background}                                     & 0.52                           & 0.92                           & 0.93                      & 0.90                      & 0.90                         & 0.57                     & 0.79                         & 0.43                        & 0.88                         & 0.86                         & 0.92                        & 0.77                        \\
MBS \cite{sajid2017universal}                                          & 0.52                           & 0.92                           & 0.93                      & 0.90                      & 0.90                         & 0.57                     & 0.79                         & 0.53                        & 0.88                         & 0.86                         & 0.92                        & 0.80                        \\
BMOG \cite{martins2017bmog}                                            & 0.38                           & 0.93                           & \textbf{0.95}             & 0.84                      & \textbf{0.96}                & 0.69                     & 0.79                         & 0.69                        & 0.73                         & 0.79                         & 0.92                        & 0.78                        \\
CANDID \cite{mandal2018candid}                                         & 0.55                           & 0.92                           & 0.91                      & 0.66                      & 0.92                         & 0.81                     & 0.80                         & 0.83                        & 0.78                         & 0.87                         & 0.92                        & 0.85                        \\
SBBS \cite{varghese2017sample}                                         & 0.73                           & 0.93                           & 0.49                      & 0.94                      & 0.91                         & 0.88                     & 0.81                         & 0.45                        & 0.79                         & 0.81                         & 0.90                        & 0.74                        \\
SuBSENSE \cite{st2014subsense}                                         & 0.75                           & 0.94                           & 0.79                      & 0.69                      & 0.86                         & 0.87                     & 0.82                         & 0.80                        & 0.89                         & 0.85                         & 0.91                        & 0.86                        \\
SharedModel \cite{chen2015learning}                                    & 0.78                           & 0.94                           & 0.62                      & 0.88                      & 0.82                         & 0.89                     & 0.82                         & 0.73                        & 0.89                         & 0.91                         & 0.86                        & 0.85                        \\
CwisarDH \cite{de2014change}                                           & 0.61                           & 0.93                           & 0.94                      & 0.84                      & 0.90                         & 0.75                     & 0.83                         & 0.32                        & 0.75                         & 0.91                         & 0.77                        & 0.68                        \\
WisenetMD \cite{lee2019wisenetmd}                                      & 0.75                           & \textbf{0.95}                  & 0.87                      & 0.71                      & 0.87                         & 0.87                     & 0.84                         & 0.80                        & 0.89                         & 0.85                         & 0.91                        & 0.86                        \\
AMBER \cite{wang2014fast}                                              & 0.77                           & 0.93                           & 0.93                      & 0.85                      & 0.95                         & 0.63                     & 0.84                         & 0.65                        & 0.72                         & 0.79                         & 0.91                        & 0.77                        \\
CwisarDRP \cite{de2017wisardrp}                                        & 0.69                           & 0.92                           & 0.91                      & 0.84                      & 0.92                         & 0.82                     & 0.85                         & 0.71                        & 0.80                         & 0.91                         & 0.78                        & 0.80                        \\
CVABS \cite{icsik2019cvabs}                                            & 0.77                           & 0.94                           & 0.88                      & 0.81                      & 0.86                         & 0.91                     & 0.86                         & 0.83                        & 0.84                         & 0.87                         & 0.89                        & 0.86                        \\
SWCD \cite{isik2018swcd}                                               & 0.76                           & 0.93                           & 0.92                      & 0.85                      & 0.85                         & 0.88                     & 0.86                         & 0.78                        & 0.83                         & 0.82                         & 0.86                        & 0.82                        \\
DBSGen \cite{bahri2022dynamic}                                         & 0.73                           & 0.80                           & 0.90                      & 0.91                      & 0.87                         & 0.93            & 0.86                         & 0.82                        & 0.76                         & 0.80                         & 0.86                        & 0.81                        \\
FTSG \cite{wang2014static}                                             & \textbf{0.81}                  & \textbf{0.95}                  & 0.69                      & \textbf{0.95}             & 0.94                         & 0.93            & 0.88                         & 0.71                        & 0.82                         & 0.85                         & 0.91                        & 0.82                        \\
PAWCS \cite{st2015self}                                                & 0.78                           & 0.94                           & 0.94                      & 0.84                      & \textbf{0.96}                & 0.91                     & 0.89                         & 0.75                        & 0.77                         & 0.84                         & 0.90                        & 0.82                        \\
Our Method                         & \textbf{0.81}                           & 0.94                           & 0.92                      & 0.94                      & 0.91                         & \textbf{0.94}            & \textbf{0.91}                & 0.83                        & 0.89                         & \textbf{0.92}                & 0.93                        & 0.89                        \\ \hline
\end{tabular}
\end{table*}

\begin{table}[htbp!]
\caption{Performance comparison with LDB weakly supervised method \cite{zhang2022learning}, evaluated on three dynamic background videos of \textit{I2R} dataset, in terms of F-measure. The best performance achieved, in each column, is shown in bold.}
\label{table:I2R_results}
\begin{tabular}{l|rrrr}
Methods                                                           & \multicolumn{1}{l}{Campus} & \multicolumn{1}{l}{Fountain} & \multicolumn{1}{l}{WaterSurface} & \multicolumn{1}{l}{Avg} \\ \hline
LDB \cite{zhang2022learning} & \textbf{0.85}              & 0.85                         & 0.94                             & 0.88                    \\
Our Method                                                        & 0.83                       & \textbf{0.86}                & \textbf{0.95}                    & 0.88                   
\end{tabular}
\end{table}

\begin{table*}[htbp!]
\caption{Performance comparison wit the Weakly Supervised method proposed in \cite{minematsu2019simple}, evaluated on eight categories of \textit{CDnet 2014} , in terms of F-measure. The best performance achieved, in each column, is shown in bold. }
\label{table:comparasion2weakly2}
\begin{tabular}{l|rrrrrrrrr}
Methods                                        & \multicolumn{1}{l}{BadWeather} & \multicolumn{1}{l}{Baseline} & \multicolumn{1}{l}{CameraJitter} & \multicolumn{1}{l}{DynamicBg.} & \multicolumn{1}{l}{NightVideos} & \multicolumn{1}{l}{Shadow} & \multicolumn{1}{l}{Thermal} & \multicolumn{1}{l}{Turbulence} & \multicolumn{1}{l}{Avg} \\ \hline
Weak-Supervised method \cite{minematsu2019simple} & 0.72                           & \textbf{0.97}                & \textbf{0.61}                    & 0.82                                  & \textbf{0.38}                   & 0.56                       & 0.66                        & 0.58                             & 0.66                    \\
Our Method                                     & \textbf{0.89}                  & 0.92                         & 0.42                             & \textbf{0.91}                         & 0.29                            & \textbf{0.92}              & \textbf{0.78}               & \textbf{0.59}                    & \textbf{0.72}          
\end{tabular}
\end{table*}

\section{Conclusion} \label{sec:Conclusion}
In this paper, we presented a novel weakly supervised realtime method for dynamic background subtraction, which utilizes two neural networks: an autoencoder for static background image generation and a U-Net for dynamic background image generation. While the autoencoder learns in an unsupervised manner, the U-Net requires pixel-wise ground-truth labels for supervised training. However, obtaining pixel-wise annotations can be an expensive and time-consuming task. To overcome this challenge, we prepared these labels in a weakly supervised way by selecting training frames that do not contain any moving objects. The autoencoder can generate the static background image by leveraging the temporal correlation between frames. After performing background subtraction and thresholding, the resulting image represents the dynamic background since the input image is moving object-free and only contains dynamic and static background. The U-Net then trains with the same moving object-free sequence of images and the binary dynamic background images as the ground-truth labels. During testing, we can feed an input image to the networks and obtain the static and dynamic background images in the output, resulting in a clean foreground image without dynamic background motions.

Our experiments on various sequences demonstrated that our method is effective in real-world scenarios. Our algorithm outperformed all top-ranked unsupervised methods as well as a weakly supervised method. We performed equally to another state-of-the-art weakly supervised method \cite{zhang2022learning}, which is specifically designed for handling dynamic background scenes.

In summary, our proposed method has a training phase followed by an online test phase, during which it can effectively detect dynamic background artifacts and separate them from the moving object foreground.

For future work, we plan to incorporate data augmentation techniques to acquire more comprehensive training data that includes pixel-wise dynamic background labels for images containing moving objects. Additionally, we aim to incorporate data augmentation with different brightness levels to handle illumination changes effectively.

\bibliographystyle{IEEEtran}
\bibliography{ref}
\end{document}